\documentclass[runningheads]{llncs}
\usepackage[T1]{fontenc}
\usepackage{tikz}
\usepackage{amsmath} 
\usepackage{amssymb} 

\usepackage{verbatim}
\usepackage{mathtools}
\usepackage{subcaption}

\usepackage{pgfplots}
\usepackage{multirow}

\usepackage{graphicx}
\begin{document}

\title{Attention, Filling in The Gaps for Generalization in Routing Problems}

\author{
Ahmad Bdeir\inst{1} \and
Jonas K. Falkner\inst{2} \and
Lars Schmidt-Thieme\inst{3}
}
\authorrunning{Bdeir et al.}
%
\institute{
Hildesheim Universität, Hildesheim, Germany\\
\email{bdeira@uni-hildesheim.de\inst{1}, falkner@ismll.de\inst{2}, schmidt-thieme@ismll.de\inst{3}}
}
\maketitle              
\begin{abstract}
Machine Learning (ML) methods have become a useful tool for tackling
vehicle routing problems, either in combination with popular heuristics or as
standalone models. However, current methods suffer from poor generalization when tackling problems of different sizes or different distributions. As
a result, ML in vehicle routing has witnessed an expansion phase with new methodologies being created for particular problem instances that become infeasible at larger problem sizes.

This paper aims at encouraging the consolidation of the field through understanding and improving current existing models, namely the attention model by Kool et al. We identify two discrepancy categories for VRP generalization. The first is based on the differences that are inherent to the problems themselves, and the second relates to architectural weaknesses that limit the model's ability to generalize. Our contribution becomes threefold: We first target model discrepancies by adapting the Kool et al. method and its loss function for Sparse Dynamic Attention based on the alpha-entmax activation. We then target inherent differences through the use of a mixed instance training method that has been shown to outperform single instance training in certain scenarios. Finally, we introduce a framework for inference level data augmentation that improves performance by leveraging the model's lack of invariance to rotation and dilation changes. 

\end{abstract}

\section{Introduction}
The vehicle routing problem, first introduced in \cite{dantzig_truck_1959}, lends itself as one of the most studied combinatorial optimization problems in the field. It describes the process of optimally serving a set of customers with fixed demands using a fleet of vehicles from a fixed depot. The attention on VRPs has only been bolstered by the more recent explosion of practical applications in the production and delivery of goods whether from local or global sources. This is especially amplified by the global COVID pandemic that forced brick and mortar retail shops to shut down for extended periods of time.  

Current approaches to tackling VRPs are divided between classical heuristic methods, machine learning, or a combination of the two. Exact methods do exist for the simpler problem variants however they become infeasible when tackling larger problem instances due to computational constraints \cite{bai_analytics_2021}. Classical heuristics also suffer from a similar constraint though to a lesser extent. Combing through the solution space becomes very expensive, especially with larger neighborhoods. Additionally, classical heuristics produce a solution for every problem instance independently. No general model is learned and every further instance will require a reset of the solving time. More time-constrained VRP applications have then come to rely on machine learning to learn a generally applicable model for similar problem types.
Despite this trend however, the research has so far been focused on improving performance in smaller problem instances. Scaling the models to even slightly greater sizes has proven inefficient in terms of the training time required and the incurred computation cost. As a result, when targeting these larger VRPs, most methods rely on breaking down the problem into a set of smaller sub-problems and solving them individually. The model would then need another component to find the optimal route combination method. This is with the earlier approach beginning to gain more traction in the research field \cite{bai_analytics_2021}. 

It also seems that the issue of generalization remains somewhat overlooked. There is a rush towards expanding the field with new methods that improve results on the default small instances rather than consolidation of existing research and scaling improvements. This paper will attempt the latter where the second option becomes selecting promising current methods and building on possible flaws so that they can generalize better to large instances. Specifically, we aim at improving one such model by Kool et al. that is based on a combination of reinforcement learning and self-attention \cite{kool_attention_2019}. The paper is currently used a base for multiple other models and we hope the improvements derived would extend to those variations as well. Our contribution becomes as follows:
\begin{itemize}
\item Identification, classification, and attempted resolution of the model and problem inconsistencies that hinder generalization. Through the proposed solutions we show that the identified issues are valid and we lay the grounds for further research
\item Creation of an adapted attention model that improves performance in both upscaling and downscaling scenarios 
\item Proposition of a modified REINFORCE loss for sparse attention activation functions ($\alpha-entmax$) and a modified VRP training scheme that reduces training time and increases performance
\item Proposition of an inference-stage data augmentation method that boosts both regular performance and generalization ability for most ML-based VRP   construction methods.
\end{itemize} 

\section{Related Work}

 The attention model by kool et al. leverages the transformer model and replaces RNN based structures with attention in the encoder and the decoder. The encoder takes node-wise features and applies a linear projection then updates the resulting embeddings through N attention layers. This is done once at the beginning of the training process. The decoder uses these embeddings as the keys and the values along with a context vector as queries in order to determine the best node to add to the route. The context vector represents the current state of the route and so the problem is solved sequentially, adding one node at a time till completion \cite{kool_attention_2019}.  

 Wu et al. combine ML and classical heuristics to achieve state-of-the-art results within a smaller time frame than the heuristics alone \cite{wu_learning_2021}. An initial solution is first generated using nearest insertion, the improvement problem is then formulated into a reinforcement learning process. The state used is the current solution, the action is a node pair in the solution to perform a local operator on, and the reward is the difference between the current solution length and the newly generated one. The goal of the model is then to learn an optimal policy for selecting the best local operator pairs. Wu et al. also limit the maximum number of actions that can be performed per iteration to $T$. It has been shown that larger Ts lead to better results but slower time performance and as such a balance is to be made based on the required performance and the training time constraints \cite{wu_learning_2021}. 

To augment the input node features, Wu et al. use self-attention on the encoder and decoder level similar to kool et al \cite{wu_learning_2021}. Both of these models were trained on a maximum problem graph size of 100. This is mainly because the attention model encoding is very computationally expensive at higher sizes. The Wu et al. model is then infeasible when running on the large-scale problems proposed by Arnold et al. even when using smaller T values. In addition, the generalization studies performed by the paper show a drastic increase in the optimality gap when the model is used on different problem size instances. This was true for both downscaling and upscaling respectively \cite{wu_learning_2021,kool_attention_2019}. Attention models become unable to tackle large, real-world problem instances.

Another method based on the Kool et al. attention model is Policy Optimization with Multiple Optima (POMO). In their paper, Kwon et al. dissect the solution construction problem into the selection of a first node $\pi_1 = p_\theta(a_1|s_1)$ and the selection of the remaining nodes $\pi_t = p_\theta(a_t|s_t, a_{1:t-1})$ where $t \in /{2,...,T/}$. When using this formulation the final solution becomes contingent on the first action $a_1$ similar to a form of bias \cite{kwon_pomo_2021}. The POMO model exploits this lack of consistency to introduce a low variance baseline for REINFORCE applications in VRPs. The model first samples a set of $N$ different nodes as "favorable" starting actions ${a_1^1,..., a_1^N}$ and constructs N different trajectories in parallel. The mean cost of N trajectories is used as the baseline when updating the model weights with the REINFORCE algorithm. Kwon et al. show this achieves lower variance and enhances the model's overall performance drastically\cite{kwon_pomo_2021}. 

Peng et al. criticize another limitation to the attention-based models and present the Adaptive Dynamic Attention for VRPs(ADM-VRP) as a possible solution. They claim that the dynamic nature of the problem is poorly represented in the original Kool et al. model. The graph embedding is only calculated once at the beginning of the solving process {\cite{peng_deep_2020}} and is not updated to reflect the changes in the remaining unserved nodes. To tackle this, ADM instead recomputes the embeddings after every partial solution. The decoder treats the remaining customers as a separate problem w.r.t. the original model and utilizes the encoding of the new subgraph when making decisions. The authors show that ADM performs better on same size problems, and is also able to generalize much better on instances of different sizes {\cite{peng_deep_2020}}. 

\section{Preliminaries}

\subsection{Problem Definition}
We define a CVRP instance of size $N$ over an undirected graph $G(V,E)$ where $V=\{v_0,...,v_N\}$ is the set of vertices and $E =\{ e_{ij}=(v_i,v_j):v_i,v_j\in V, i<j\}$ is the set of edges connecting the vertices. We also define the symmetric matrix $C = [c_{ij}]$ that corresponds to the cost of traversing edge $(e_{ij}$ as the travel distance between the graph nodes. We fix the node $v_0$ to the depot node which holds a homogeneous fleet of $K$ vehicles with a carrying capacity of $D^N$. The goal is to serve the set of customers $V \text{\textbackslash} \{v_0\}$ with individual non-negative demands $d_i>0$ while minimizing the total travel cost incurred. Each node, except for the depot, is visited exactly once by one vehicle.

The solution for a CVRP instance or tour can then be defined as the sequence $\tau = \{(\tau_0,...,\tau_T)ß$ where $T$ is the number of individual routes traversed and $\tau_t$ is a subset of the graph and begins and ends with the depot node $\tau_t = \{ v_0, \subset V \text{\textbackslash} \{v_0\}, v_0\}$. For the CVRPs tackled in this paper, we assume an unconstrained number of vehicles $K$ and as such, an unconstrained number of tours $T$. 

\subsection{Original Model}

The original Kool et al. model is based on the transformer architecture with an attention encoder-decoder. The main difference is the lack of positional encoding as the input order of the nodes has no significance for the problem representation\cite{kool_attention_2019}.

\subsubsection{Encoder}

The encoder first calculates an initial $d_h$-dimensional graph node embedding through a learned projection: 
\begin{equation}
h_i^{(0)} = 
\begin{cases}
W_xx_i + b_x & \text{if}\ i \neq 0 \\
W_0x_i + b_0 & \text{if}\ i = 0 \\
\end{cases}
\end{equation}
where $x_i$ is the $d_x$-dimensional node features ($d_x = 3$ for CVRP) and separate weights are used for the depot embedding. We then update these embeddings with $N=3$ attention layers to compute the final embeddings $h_i^{(N)}$. Here an attention layer is defined as an MHA sublayer and a fully connected feed-forward sub-layer (FF).
The MHA layer is used as the message passing algorithm in the graph. It is the standard MHA used in the transformer model with 8 heads. As for the FF layer we use a hidden dimension of 512 and ReLU activations. Both layers also use skip connections and batch normalization:
\begin{equation}
\hat{h}_i^{(l)} = BN^l(h_i^{(l-1)} + MHA_i^{(l)}(h_1^{(l-1)},...,h_n^{(l-1)}))
\end{equation}
\begin{equation}
FF(\hat{h}_i^{(l)}) = W_1^FReLU(W_0^F\hat{h}_i^{(l)}+b_0^F)+b_1^F
\end{equation}
\begin{equation}
h_i^{(l)} = BN^l(\hat{h}_i + FF^l(\hat{h}_i))
\end{equation} 

This gives us the final node embeddings $h_i^{(N)}$. The encoding is done once for the entire solving process and the embeddings are reused statically for every decoding step. We also calculate a graph embedding $\bar{h}$ as an aggregation of the total node embeddings $\bar{h} = \dfrac{1}{n}\sum_{i=1}^n{h_i^{(N)}}$ to be used for the decoder. 

\subsubsection{Decoder}

The problem is solved sequentially with a node being visited at every construction step $t \in \{1,…, T\}$. The model uses a context vector $h_c$ and the node embeddings to create a probability distribution over the remaining nodes and sample the next action. The theory behind using a context vector $h_c$ is guiding the decoding process under the current problem state. To calculate $h_c$ we first generate an initial vector $h’_c$ as:
\begin{equation}
h_{(c)}^{(N)} =    
\begin{cases}
[\bar{h}^{(N)}, h^{(N)}_{\pi_{t-1}}, \hat{D}_t], & \text{if}\ t>1 \\
[\bar{h}^{(N)}, h^{(N)}_{0}, \hat{D}_t], & \text{otherwise}
\end{cases}
\end{equation} 
where $\bar{h}^{(N)}$ is the average graph embedding, $h^{(N)}_{\pi_{t-1}}$ is the last visited node, $\hat{D}_t$ is the remaining vehicle capacity and [.,.,.] is used as the concatenation operator.

 We then pass $h’_c$ through a single M-head attention layer) to get the final context vector $h_c$.  The parameters are not shared with the encoder layers, and only a single query $q_c$, the linear transformed $h’_c$ vector, is computed for every head. This gives:

\begin{equation}
q_{(c)}^m = W_q^mh_c', \;\; k_{j}^m = W_k^mh_j^N, \;\; v_{j}^m = W_v^mh_j^N
\end{equation}
and the remaining MHA operations are done as discussed in the transformer section. Finally, a single head attention layer is used to calculate probabilities $p(\pi_t|X,\pi_{1:t-1})$:

\begin{equation}
q = W_qh_c, \;\; k = W_kh_j^N
\end{equation}

\begin{equation}
u_{j}=
\begin{cases}
\text{C . tanh}(\dfrac{q_i^Tk_j}{\sqrt{d_k}}) , & \text{if}\ d_j <  \hat{D}_t \text{ and } x_j \notin \pi_{1:t-1} \\
-\infty, & \text{otherwise}
\end{cases}
\end{equation}

where $C$ is the clipping operator between $[-10,10]$. This gives:
\begin{equation}
\label{attention-output}
p_\theta(\pi_t=x_j|X,\pi_{1:t-1}) = \text{softmax}(u_j)
\end{equation} 
For training, the model samples the next action from the calculated distribution and for inference, it takes a greedy approach and select the node with the highest probability.

\section{Targeting Generalization}

In order to alleviate the issues with generalization, we first divide the discrepancies between problems of different graph sizes into fixed problem differences and model differences. Fixed differences are discrepancies that are inherent to the problem itself and cannot  be changed by altering the model. These include differences in capacities, differences in the action space, differences in the required number of routes, etc. These cannot be changed and instead the model must be altered or trained to accommodate them. As for model differences, they are related to the model's ability to represent the problem properly. The following section discusses the differences tackled in this paper.
\subsection{Inherent Differences}

In the case of inherent issues, we identify the following key problems:
\begin{itemize}
    
\item \textbf{Node Density:} Training data is sampled in the unit square by default. Any increase in the number of nodes generated in the same area will cause an increase in overall node density. This changes the typically expected distances between the nodes and can lead to model confusion when selecting the next best action.  

\item \textbf{Capacity Difference:} Larger problem instances utilize vehicles with greater capacities that can carry more load. When faced with a similar distribution in demands, that leads to a change in the average node-wise route length. We theorize that the model could be learning the average route length and might tend to make longer or shorter routes depending on the problem that it is generalizing to.  
 
 \end{itemize}
\subsection{Model Differences}
 As for the model differences, we identify:
 \begin{itemize}

\item \textbf{Attention Dilution:} By default, the attention mechanism uses the softmax activation function which is unable to deliver 0 attention to any node \cite{kool_attention_2019,peters_sparse_2019}. By increasing or decreasing the number of nodes we then effectively concentrate or dilute attention and present the model with attention distributions that are unfamiliar to it. 

\item \textbf{Static Encodings:} The Kool et al. model is static in the embedding technique, the encodings are calculated once in the beginning and reused for every subsequent decoding step \cite{kool_attention_2019}. It is unable to capture the dynamic changes in the problem as the solution develops \cite{peng_deep_2020}. The nodes that have already been visited in a previous route are no longer relevant to the selection of the next node in the current route. Available attention, embedding capacity, and model resources are exhausted.  
\end{itemize}

We realize there are other architectural decisions that could be examined but we recognize the above as more major issues and tackle them in specific. 

\section{Methodology}

\subsection{Dynamic Encoder}

Kool et al. encode the graph once at the 0-time step and then reuse the encodings in every decoding step until termination \cite{kool_attention_2019}. In their paper, Peng et al. state that after a route in the tour is generated, the remaining unvisited nodes form a new subproblem \cite{peng_deep_2020}. The static embeddings that were previously calculated become less suited to represent the new structure information. We follow Peng et al. in introducing more frequent encodings to resolve this. Specifically, we re-encode the remaining node features every time a partial solution is found. Formally, the embedding of the problem node $i$ then becomes:
\begin{equation}
h_{(i)}^{(t)} =    
\begin{cases}
ME(h_0^0,...,h_n^0), & \text{if}\ \pi_{t-1} = x_{(depot)} \\
h_{(i)}^{(t-1)}, & \text{otherwise}
\end{cases}
\end{equation} 

Re-embedding the new subproblem simulates training on smaller graph sizes which would typically help to downscale. However, even for the task of upscaling, breaking down the problem allows it to eventually reach familiar sizes that resemble the training data. This is similar to approaches that rely on the partitioning of the complete problem into smaller subgraphs. The difference here is that problem partitions are done sequentially after a solution route is established. Another difference is that the new problem graph is a subset of the older graph node-set. This is opposed to normal graph partitioning methods where the subproblems are independent, determined in the beginning, and can be solved in parallel. 

\begin{figure}
    \centering
     \scalebox{.7}{\input{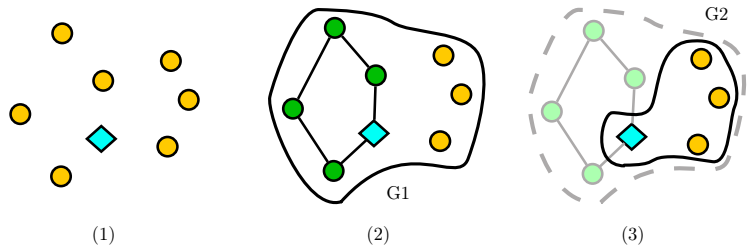}}
    \caption{Problem graph re-encoding at every partial solution found. The new instance G2 is a subgraph of G1 and is encoded as a separate problem.}
\end{figure}

For the model, we follow Peng et al. in their implementation of the re-encoding logic. During the training process, the encoder can only be run for the entire batch, this includes problem instances that have not yet finished their current routes. The model can perform the computation and discard the values but that would be very wasteful in terms of computation resources. To resolve this, a problem instance that has already reached the depot is forced to remain at the depot until all other instances in the batch also finish their partial solutions. 
\begin{equation}
\pi^b_t =    
\begin{cases}
x_{depot}, & \text{if}\ \pi^b_{t-1} = 0 \text{ and }  \sum_{b=0}^{B}{\pi^b_{t-1}} = 0\\
DECODE, & \text{otherwise}
\end{cases}
\end{equation} 
where $B$ is the batch size and $b \in B$.

\subsection{$\alpha$-entmax Implementation}

We recognize two options for the implementation of the $\alpha$-entmax activation in the place of the softmax activations. This is based on the two roles the softmax plays in the attention model. Internally, the softmax function is used to compute the normalized attention weights in the encoder and the decoder \cite{vaswani_attention_2017}. Applying the entmax function here is straightforward and it is a simple replacement. This has been tried in NLPs and has shown to help remove noise when processing the attention weights by removing the effect of irrelevant data points, see \cite{peters_sparse_2019}. 

However, the softmax is also used in order to map the final attention score to a probability distribution for node selection \cite{kool_attention_2019}. Introducing the entmax function here is more complicated with the use of the REINFORCE loss. Applying the entmax function without any changes heavily degrades performance. We find that the model converges early to relatively bad actions. We suspect that the high sparsity further exacerbated this by encouraging these overconfident actions and assigning zero values to good actions causing them to be completely ignored early on. Their attention would also be diverted to the remaining actions that are assigned even larger probabilities. This highly limits exploration and the ability to recover good next nodes, which causes the performance to decrease dramatically.  

Williams et al. previously notes this problem of early convergence and suggests adding an entropy regularization factor in \cite{williams_function_1991} (entropy maximization). In their paper, Peters et al. also comment on a similar issue with models based on the $\alpha$-entmax activation. They propose a new loss function to replace the Negative Log-Likelihood Loss (NLL) typically used with the softmax activations. The loss incorporates the Tsallis entropy specific to each $\alpha$ value. They state that "harder" time steps that allow for multiple optimal or close to optimal actions will then lead to a higher entropy that forces the algorithm to better explore the state space \cite{peters_sparse_2019}. We follow both papers and account for the respective expected entropy for every $\alpha$. The loss becomes:
\begin{equation}
\label{final_entropy_rein}
 \nabla \mathbb{L}(\theta|s) = \mathbb{E}_{p_\theta(\pi|s)}[(L(\pi) - b(s))\nabla log \: p(\pi|s) + \beta\nabla\mathcal{H}^\pi)]
\end{equation}
where $\beta$ is a hyperparameter to control the entropy regularization amount and $\mathcal{H}^\pi$ is the corresponding entropy for the entmax activation used in policy $\pi$. This method allows for the model to learn without any performance issues or instability.

\subsection{Mixed Problem Sizes}

So far we attempt to address the generalization issue based on the observable heterogeneity between problems of different graph sizes. However, given the black-box nature of deep learning, the reasoning utilized by the model while tackling the different sizes is still ambiguous. We acknowledge this and instead attempt to leverage the model's ability to learn on mixed data. 

To do this we generate the training set $F = {f_1, ..., f_M}$ where $f_m$ is a generated data subset for a particular problem size, and $M$ is the number of different problem sizes $S$ to train on. All subsets are equal in size. At every training iteration $i$ in an epoch, the model samples a batch $b_i$ from $f_m$ where $m = (i+1) - \lfloor {i}/{M}\rfloor M$ and feeds it into the model. The different problem sizes may incur different cost magnitudes, and seeing as the objective is minimizing the cost, this could cause confusion while training. One option would be shifting the cost formula for the problem to account for the average expected problem cost.
\begin{equation}
\label{cost2}
cost_{norm}(\tau) = \sum_{t=0}^T {\dfrac{cost(\tau_t)}{s_m}}
\end{equation}
where $s_m \in S$ is the size of the problems in the current batch sample set $f_m$. However, we found no benefit from this normalization and instead used the default cost formula.

\subsection{Inference Data Augmentation}
Kwon et al. introduce the concept of instance augmentation for VRP graph data. They state that the attention-based model by Kool et al. arrives at a different solution when reformulating the same problem through small linear transformations \cite{kwon_pomo_2021}. The POMO model does this by shifting the graphs a certain amount in 8 different directions before inference. Since the relative position of graph nodes with respect to each other is conserved, solutions derived from any of the 8 instances are considered valid on the original problem. 

We take this a step further by introducing two new types of instance augmentation, graph dilation, and graph rotation. In the case of the latter the problem is rotated with respect to $(0.5,0.5)$, the center of the unit square.  Rotation degrees are determined manually however, intuitively, we only rotate the problem by multiples of 90 degrees. This avoids instances where the transformed graph nodes lie outside the unit square.
\begin{equation}
R_{O,\theta}(x_i,y_i)=
\begin{pmatrix}
\cos\theta & -\sin\theta \\
\sin\theta & \cos\theta
\end{pmatrix}
\begin{pmatrix}
x_i-a\\
y_i-b
\end{pmatrix}
+\begin{pmatrix}
a\\
b
\end{pmatrix}
\end{equation}
 where O is the center of rotation at $(a,b)$ and $\theta$ is the rotation degrees in radians. Tour cost can be calculated directly from the solution of the augmented graphs since the distances between the nodes remain constant. 

The second method, however, graph dilation, relies on scaling the distances between graph nodes with respect to a center of dilation, also taken as $O = (0.5,0.5)$. For any augmented node $i$ with coordinates $(x_i,y_i)$ the transformed coordinates become
\begin{equation}
D_{O,k}(x_i,y_i) = (k(x_i-a)+a,\;k(y_i-b)+b)
\end{equation}
 where O is the center of dilation at $(a,b)$ and $k$ is the scale factor. The inference is done on dilated graphs with different scale factors that cater towards fitting the inference problem size to the density of the training problem size. It should be noted that the costs from these solutions cannot be directly computed. Instead, we generate the solutions, apply them to the original data instances and then calculate costs. This is because the inferred cost will scale with the dilation process. 
 
 \subsection{Model Training and Evaluation}
 
For training, we follow Nazari et al. in their data generation method \cite{nazari_reinforcement_2018}. The depot node and $n$ customer node coordinates are sampled uniformly in the unit square $[0,1]$. The demands $\delta_i$ are sampled uniformly in the interval ${1,...,9}$ and normalized by the problem vehicle capacity $D^n$. This gives $\hat{\delta_i} = \delta_i/D^n$, where ${D^{20} = 30}$, ${D^{50} = 40}$ and ${D^{>100} = 50}$. 

For testing as well, we follow Nazari et al. in their use of the optimality gap
with the best-known solution \cite{nazari_reinforcement_2018}. The test dataset is also generated based on the distribution above as in \cite{kool_attention_2019,peng_deep_2020,kwon_pomo_2021,nazari_reinforcement_2018}. Given a large enough test set, this should ensure the ability of reusing the numbers as reported by other papers that also utilize this. We then circumvent the difficulties with typical benchmark evaluations that include a lack of available code, or a lack of trained models (that in turn could lead to deterioration of benchmark performance due to an unknown initialization of benchmark hyperparameters when retraining).

In terms of model time consumption, we rely on the original time values reported by every model’s corresponding authors. However, we note that it is difficult to extract meaningful insight from the comparison of these run times. The hardware configurations used in every paper are greatly varied and play a very large role in the overall time consumption. This problem is only exacerbated when models such as LKH-3 are mainly CPU reliant and not only GPU reliant. We find that the field could benefit from a unified routing library that compiles existing literature methodologies. This would facilitate benchmark evaluations in terms of time consumed on the same hardware and a more accurate performance comparisons.

\section{Results}

\subsection{Dynamic Training and entmax}

In the following section, we rely on the $1.5-entmax$ throughout the experiments. The $\alpha = 1.5$ value is used in all heads of the MHA layers during encoding and decoding. It has been shown that allowing for different degrees of sparsity between heads improves performance as different learned features do not necessarily share the same relation sparsity \cite{correia_adaptively_2019}. It is possible to use separate learned $\alpha$ values but this is not done in the paper. At current, the $1.5-entmax$ function has an optimized closed-form solution for both computations and gradients. Other values rely on the bisect method to find a close estimate of the outputs (the degree of accuracy depends on the number of bisections permitted). By allowing the adaptive sparsities we incur a very high computation time and resource cost. Should more efficient methods for estimation become available, we see this as a very promising approach to build upon.

The application of the entmax function happens twofold: once using the function only as an attention normalization mechanism, and once using the function for both normalization and probability distribution mapping. The losses used are vanilla REINFORCE and REINFORCE with expected entropy normalization respectively. All models are trained on the default problem size $C_{50}$ and the results are recorded in table \ref{results_total}.

\begin{table}
\centering
\caption{Performance comparison of the trained $C_{50}$ model with different entmax implementation methods (once with 1.5-entmax as attention normalization only, and once with entmax probability output as well).}
\label{results_total}
\begin{tabular}{|c|c|c|c|c|c|c|} 
\cline{2-7}
\multicolumn{1}{l|}{}         & \multicolumn{2}{c|}{20} & \multicolumn{2}{c|}{50} & \multicolumn{2}{c|}{100}  \\ 
\cline{2-7}
\multicolumn{1}{l|}{}         & Cost & Gap              & Cost  & Gap             & Cost  & Gap               \\ 
\hline
LKH-3                         & 6.14 & 0.00\%           & 10.38 & 0.00\%          & 15.65 & 0.00\%            \\ 
\hline
ADM-50                        & 6.48 & 5.54\%           & 10.78 & 3.85\%          & 16.55 & 5.75\%            \\ 
\hline
SADM - reg. only                & 6.43 & 4.72\%           & 10.86 & 4.62\%          & 16.48 & 5.30\%            \\ 
\hline
SADM - both & 6.34 & 3.26\%           & 10.74 & 3.47\%          & 16.28 & 4.03\%            \\
\hline
\end{tabular}
\end{table}

Overall we can see that the implementation of the entmax activation for sparse attention leads to a significant increase in the model's generalization ability. We note that when used in conjunction with the softmax activation, the model performance decreases w.r.t. to the original ADM paper on problems of the same training size. We suspect that despite the default loss function implemented allowing training without convergence problems in this scenario, it remains not ideal for use with the entmax activations in general. This is due to the problems stated in the methodology section. When 1.5-entmax is applied for both normalization and in the final output layer, the model is able to exceed the performance of the original ADM model in both the $C_{50}$ test set and the generalization sets. 

Based on these results, we deduce that attention dilution is a probable issue in the model encoding and decoding of the problem. The sparse adaptive dynamic model (SADM) is then used as a basis for the final model implemented. 

\subsection{Final Model} 

We first compare the $C_{50}$, $C_{100}$ and $C_{50/100}$ trained models to popular baselines. The baseline results here are reported from the original papers (the models are trained on the graph sizes corresponding to the test set graph sizes).
\begin{table}
\centering
\caption{Experimental results on CVRP}
\makebox[\linewidth]{
\begin{tabular}{|c|c|c|c|c|c|c|c|c|c|} \hline
\multirow{3}{*}{Method} & \multicolumn{9}{c|}{Problem Size} \\ \cline{2-10}
 & 20 &  &  & \multicolumn{3}{c|}{50} & \multicolumn{3}{c|}{100} \\ \cline{2-10}
 & Mean & Gap & Time & Mean & Gap & Time & Mean & Gap & Time \\ \hline
LKH3 & 6.14 & 0.66\% & 2h & 10.38 & 0.00\% & 7h & 15.65 & 0.00\% & 13h \\ \hline
Kool (greedy) & 6.4 & 4.23\% & 1s & 10.98 & 5.78\% & 3s & 16.8 & 7.35\% & 8s \\ \hline
Kool (sampling 1280) & 6.25 & 1.79\% & 6m & 10.62 & 2.31\% & 28m & 16.23 & 3.71\% & 2h \\ \hline
Wu et. al. (5000 impr. steps) & 6.12 & -0.33\% & 2h & 10.45 & 0.67\% & 4h & 16.03 & 2.43\% & 5h \\ \hline
POMO & 6.17 & 0.49\% & 1s & 10.49 & 1.06\% & 4s & 15.83 & 1.15\% & 19s \\ \hline
POMO + Aug. & 6.14 & 0.00\% & 5s & 10.42 & 0.39\% & 26s & 15.73 & 0.51\% & 2m \\ \hline
ADM & 6.28 & 2.28\% & 1s & 10.78 & 3.85\% & 7s & 16.4 & 4.79\% & 26s \\ \hline
SADM-50 & 6.34 & 3.26\% & 1s & 10.73 & 3.37\% & 5s & 16.28 & 4.03\% & 19s \\ \hline
SADM-100 & 6.45 & 5.05\% & 1s & 10.83 & 4.34\% & 5s & 16.23 & 3.71\% & 19s \\ \hline
SADM-Mix & 6.34 & 3.26\% & 1s & 10.75 & 3.56\% & 5s & 16.18 & 3.39\% & 19s \\ \hline
SADM-Mix + Aug. & 6.24 & 1.63\% & 10s & 10.6 & 2.12\% & 38s & 15.99 & 2.17\% & 2m \\ \hline
\end{tabular}
}
\end{table}

We see that the model shows competitive performance on all the given problem sizes despite being trained only for specific instance sizes. The model's generalization ability is also able to outperform both the Kool et al. model and the base ADM models when they are trained for the corresponding test size. This is true for both upscaling and downscaling instances. We also notice that the model trained on mixed instances of sizes 50 and 100 (SADM-Mix) seems to outperform the model trained on 100 alone (SADM-100). This is even on $C_{100}$ test instances. Further experiments with other models are required to see if the same behavior is exhibited. If so, this would prove to be a beneficial training method that saves both training time and increases model abilities.

We perform a comparison between the generalization abilities of the different attention models in table \ref{comparison-gen}. 

\begin{table}
\centering
\caption{Generalization comparison with base models trained on $C_{50}$} 
\label{comparison-gen}
\begin{tabular}{|c|c|c|c|c|} \cline{2-5}
\multicolumn{1}{c|}{\multirow{2}{*}{}} & \multicolumn{2}{c|}{20} & \multicolumn{2}{c|}{100} \\ \cline{2-5}
\multicolumn{1}{c|}{} & Cost & Gap & Cost & Gap \\ \hline
LKH-3 & 6.14 & 0.00\% & 15.65 & 0.00\% \\ \hline
Kool-50 (greedy) & 6.8 & 10.75\% & 16.96 & 8.37\% \\ \hline
Kool-50 (sampling) & 6.63 & 7.98\% & 16.34 & 4.41\% \\ \hline
ADM-50 & 6.48 & 5.54\% & 16.55 & 5.75\% \\ \hline
SADM-50 & 6.34 & 3.26\% & 16.28 & 4.03\% \\ \hline
SADM-Mix & 6.34 & 3.26\% & 16.18 & 3.39\% \\ \hline
SADM-Mix + Aug. & 6.24 & 1.63\% & 15.99 & 2.17\% \\ \hline
\end{tabular}
\end{table}

\begin{figure}[h!]
\centering
  \includegraphics[width=11cm]{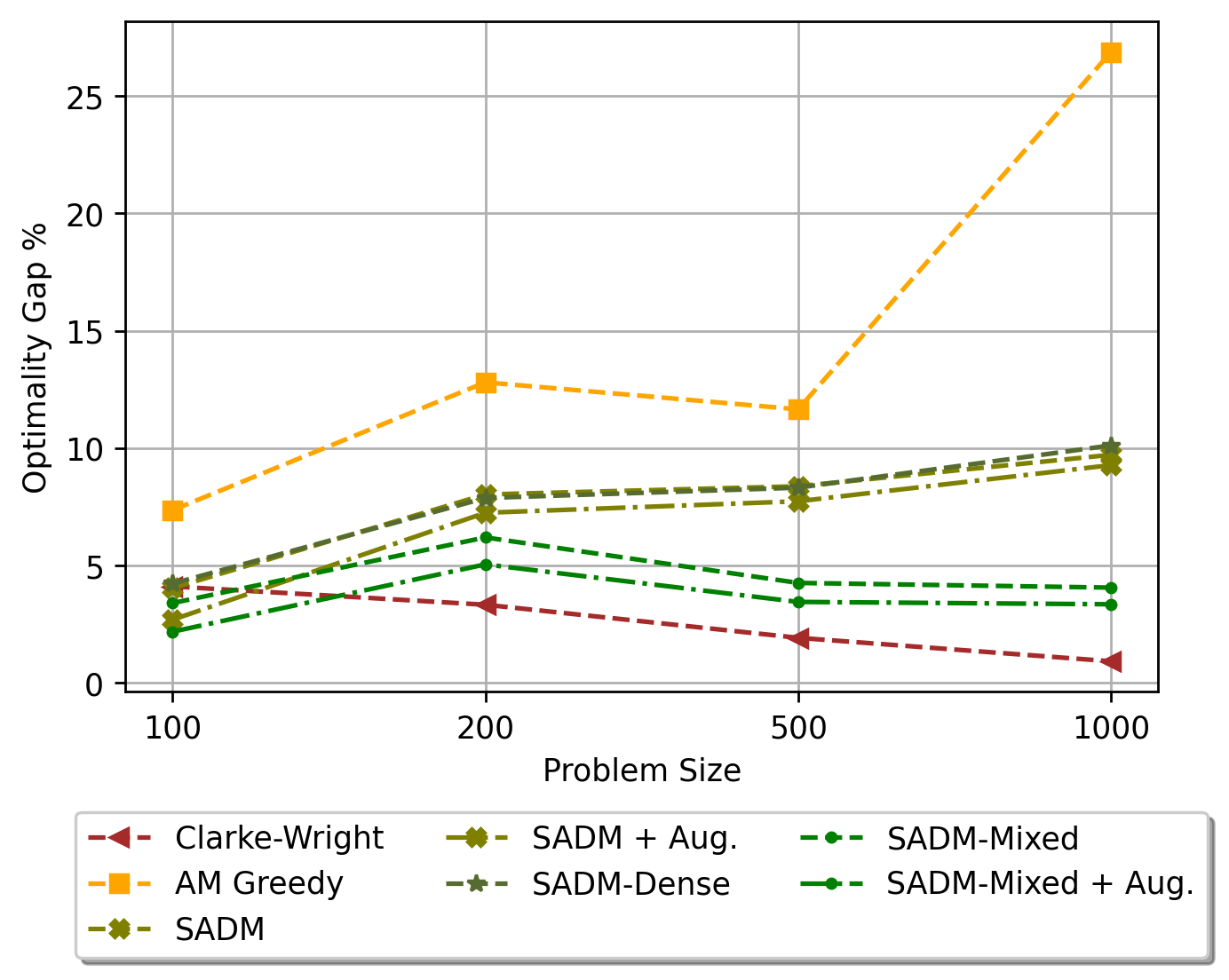}
  \caption{Comparison of the Generalization ability on large graph sizes}
  \label{ref-fig}
\end{figure}

We find that both upscaling and downscaling results have improved dramatically even when compared with the already dynamic attention model proposed by \cite{peng_deep_2020}. We extend this to compare with generalization results on significantly greater problem sizes as depicted in table \ref{big_generalize} and figure \ref{ref-fig}.

\begin{table}
\centering
\caption{Generalization ability on large graph sizes in table form}
\label{big_generalize}
\begin{tabular}{|c|c|c|c|c|c|c|} \cline{2-7}
\multicolumn{1}{c|}{} & \multicolumn{2}{c|}{200} & \multicolumn{2}{c|}{500} & \multicolumn{2}{c|}{1000} \\ \cline{2-7}
\multicolumn{1}{c|}{} & Cost & Gap & Cost & Gap & Cost & Gap \\ \hline
LKH-3 & 26.8 & 0.00\% & 61.87 & 0.00\% & 119.02 & 0.00\% \\ \hline
Clarke-Wright & 27.69 & 3.32\% & 63.1 & 1.99\% & 120.2 & 0.99\% \\ \hline
AM (100) & 30.23 & 12.80\% & 69.08 & 11.65\% & 151.01 & 26.88\% \\ \hline
SADM-50 & 28.95 & 8.02\% & 67.05 & 8.37\% & 130.58 & 9.71\% \\ \hline
SADM-100 & 28.5 & 6.34\% & 65.1 & 5.22\% & 125.66 & 5.58\% \\ \hline
SADM-Mix & 28.46 & 6.19\% & 64.5 & 4.25\% & 123.84 & 4.05\% \\ \hline
SADM-Mix + Aug & 28.15 & 5.04\% & 64 & 3.44\% & 122.99 & 3.34\% \\ \hline
\end{tabular}
\end{table}

While the model does seem to remain usable for the $C_{200}$ problem instances, the performance continues to fall off with the increasing sizes. It should be noted that not only does SADM perform better than the Kool et al. model, the falloff between $C_{500}$ and $C_{1000}$ instances is greatly reduced. For Kool et al. the optimality gap increases almost two-fold, this is contrasted with a very slight increase in the gap for the SADM model. We assume that the architectural changes implemented manage to counteract a lot of the model deficiencies with greater scaling. It should also be noted that any lower optimality gap for higher problem sizes could be the result of a fall-off in LKH-3 performance given the time constraint. 

\subsection{Graph Augmentation}

\begin{figure}[h!]
\centering
  \includegraphics[width=10cm]{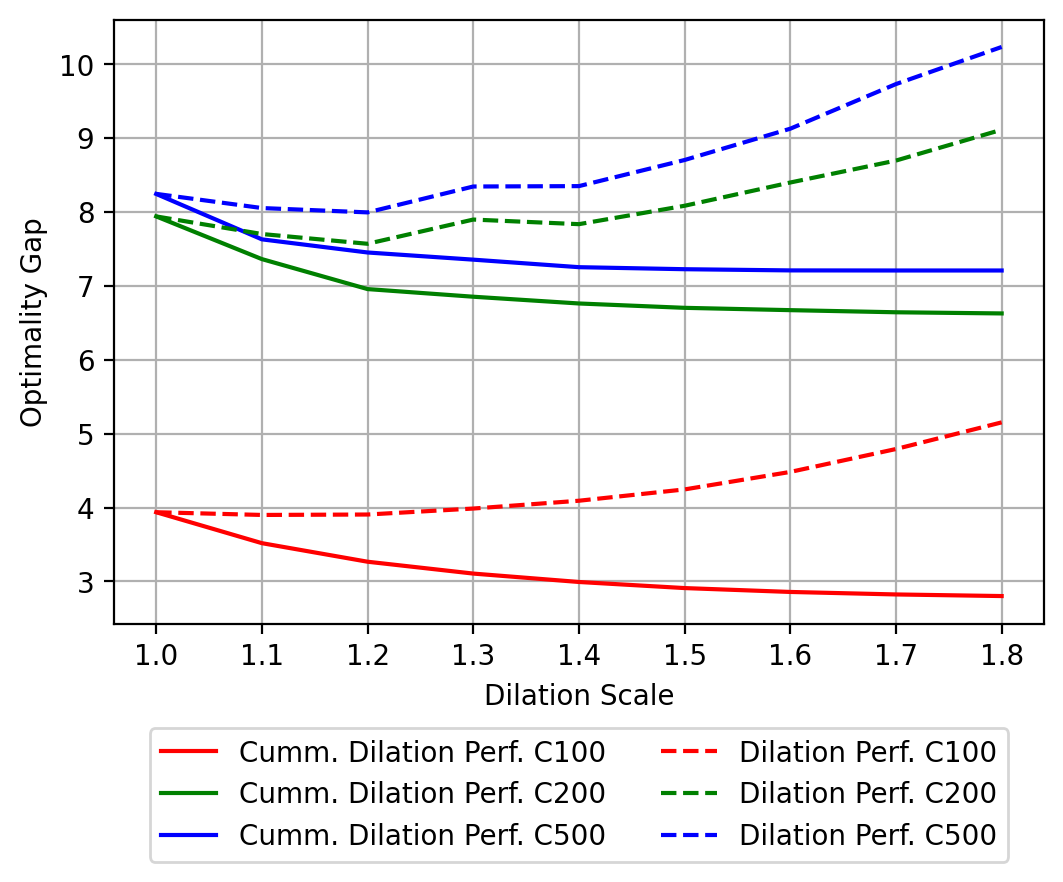}
  \caption{Model performance for $C_{100}$ through $C_{500}$ using different dilation scales}
  \label{dilate_ablation}
\end{figure}

In terms of data augmentation, we use two main methods of manipulating the inference test set. We are able to rotate the graphs or enlarge and compress them about the center of the unit square they were sampled in. We refer to the process of enlarging and compressing the graphs as dilation. Initially, dilation was only attempted as graph compression, however, an ablation study was conducted to measure any possible benefits from increasing the graph scale. Intuitively, we predicted this would cause a degradation in model performance since graph nodes will no longer be restricted to the trained unit square coordinate space. The graphs were scaled by $k \in [1,1.8]$ in intervals of 0.1. The results are recorded for the respective $C_{100-500}$ validation datasets and reported in the figure \ref{dilate_ablation}. We record the cumulative performances (take the lowest cost of each instance in the set in different dilations), and single dilation performances.

We notice that model performance continued to benefit when scaling up the graph node coordinates. This was counter to initial thoughts and could prove to be an inexpensive solution to resolving density discrepancies even when upscaling. For the final model, we rely on 5 dilation values $\{1.1, 1.2, 1.6, 1.8\}$ along with the original problem scale. In practice, should the method be applied to graph sizes with high inference time, we can adjust the dilation factors to predetermined values that best suit the target problem size.

As noted in table \ref{big_generalize}, the impact of dilation decreases as the problem size grows significantly. Even higher dilation factors are used but no benefit can be drawn which is typically the case as we can see in \ref{dilate_ablation}. We hypothesize that this is due to the very large difference between training and inference densities. In order to mimic the training density, the problem nodes must be dilated too far a distance from the original unit square. The encoder does not have the degree of flexibility required to capture the graph information accordingly at that point. The fall-off in encoder performance counteracts any benefit seen from density dilution. 

\begin{figure}[h!]
  \centering
  \includegraphics[width=9.5cm]{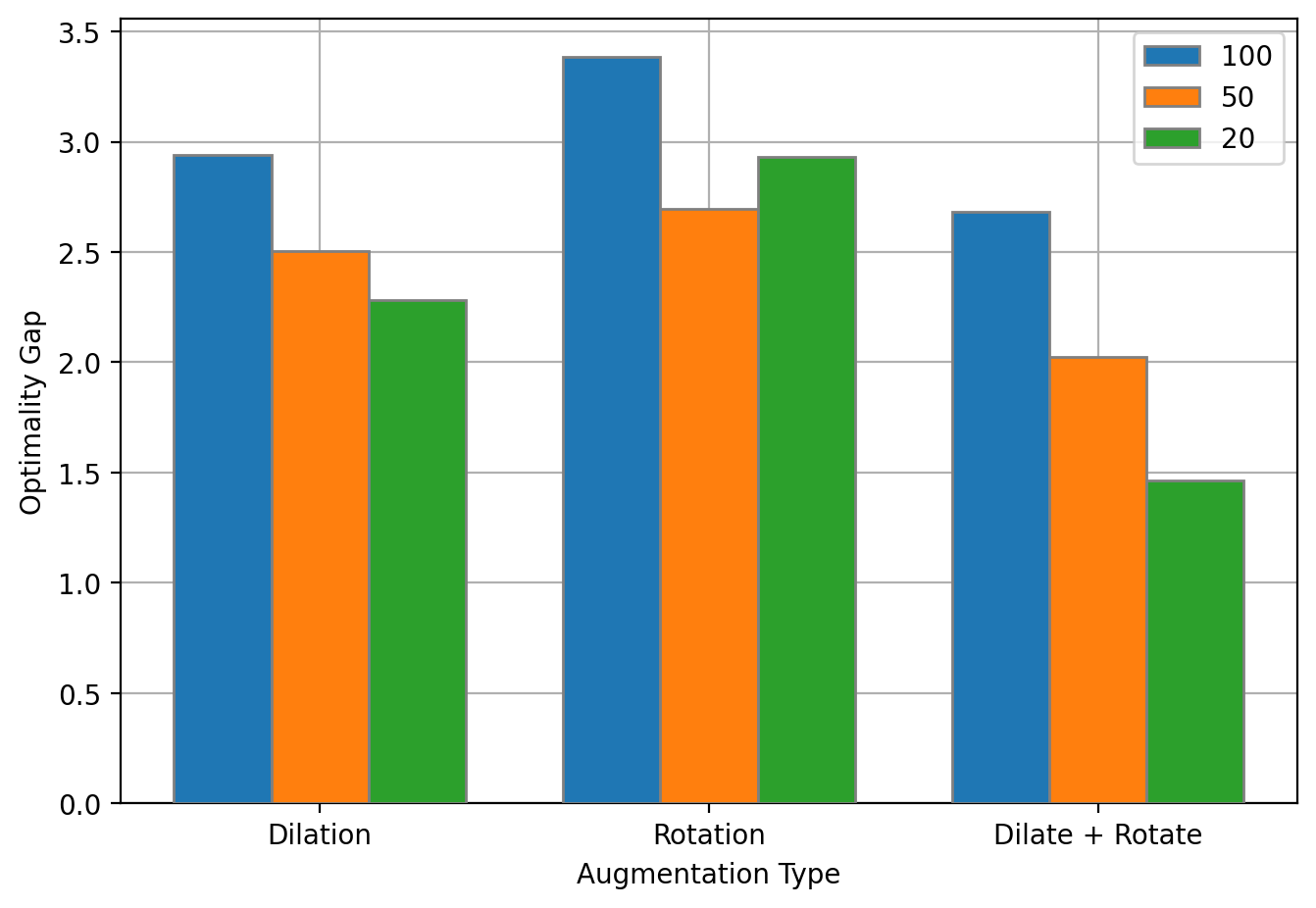}
  \caption{Model performance for $C_{20}$ through $C_{100}$ using dilation vs rotation}
  \label{dilate_rotate}
\end{figure}

We also find that the larger augmentation benefits come from the dilation operation rather than rotation as can be seen in \ref{dilate_rotate}. This is especially in generalization tasks and likely due to dilation mitigating the graph density issue. However, and with the encoder proving able to handle coordinates slightly outside the unit square, we increase the number of possible rotations from multiples of 90 to multiples of 45. The optimal values of rotation and dilation factors for each problem size should be determined to benefit inference performance while limiting inference time and the number of solving instances. It should be noted, however, that augmented inferences can be done in batches in parallel, this is opposed to the sequential nature of beam search methods. This makes solving a large number of augmentations much faster in comparison.

\section{Conclusion}

In conclusion, in this paper we provide a series of possible solutions that benefit both the upscaling and downscaling abilities of the attention model by Kool et al. This also inherently verifies our assumptions on the underlying causes of poor performance when moving to other graph sizes. These problems are alleviated but not solved completely as can be seen in the results. The model does perform significantly better than the base attention model and other baselines but the optimality gap still grows as we generalize farther away from the original training set size. 

Overall the solutions provided set up a framework for both future works on generalization as well as attention-based models for VRPs in general. Implementing sparse attention, specifically using the $\alpha$-entmax activation, improves performance at a very small computation cost. This is the same with regard to mixed training instances and the proposed inference level data augmentation. More research is required on the topic but we believe these are good first steps in the direction of expanding pure ML methods in solving the CVRP. 

\subsubsection{Acknowledgements} 
This work was supported by the German Federal Ministry of Education and Research (BMBF), project "Learning to Optimize" (01IS20013A:L2O)

%
%
%
%

\end{document}